\documentclass[journal]{IEEEtran}
\usepackage{amsmath,amsfonts}
\usepackage{algorithmic}
\usepackage{algorithm}
\usepackage{array}
\usepackage[caption=false,font=normalsize,labelfont=sf,textfont=sf]{subfig}
\usepackage{textcomp}
\usepackage{stfloats}
\usepackage{url}
\usepackage{verbatim}
\usepackage{graphicx}
\usepackage{cite}
\usepackage{hyperref}
\title{The dangers in algorithms learning humans' values and irrationalities}
\author{Rebecca Gorman, Stuart Armstrong
\thanks{Rebecca Gorman is with the Berkeley Existential Risk Initiative and Aligned AI}
\thanks{Stuart Armstrong is with the Future of Humanity Institute, Oxford University, and with MIRI, the Machine Intelligence Research Institute.}
}

\begin{document}

\maketitle

\begin{abstract}
For an artificial intelligence (AI) to be aligned with human values (or human preferences), it must first learn those values. AI systems that are trained on human behavior, risk miscategorising human irrationalities as human values -- and then optimising for these irrationalities. Simply learning human values still carries risks: AI learning them will inevitably also gain information on human irrationalities and human behaviour/policy. Both of these can be dangerous: knowing human policy allows an AI to become generically more powerful (whether it is partially aligned or not aligned at all), while learning human irrationalities allows it to exploit humans without needing to provide value in return. This paper analyses the danger in developing artificial intelligence that learns about human irrationalities and human policy, and constructs a model recommendation system with various levels of information about human biases, human policy, and human values. It concludes that, whatever the power and knowledge of the AI, it is more dangerous for it to know human irrationalities than human values. Thus it is better for the AI to learn human values directly, rather than learning human biases and then deducing values from behaviour.
\end{abstract}
\begin{IEEEkeywords}
Alignment, AI safety, Value-learning, Preferences, Recommender Systems
\end{IEEEkeywords}

This work has been submitted to the IEEE for possible publication. Copyright may be transferred without notice, after which this version may no longer be accessible.

\section{Introduction}

Algorithms are becoming more powerful, and might in future become so powerful that they form superintelligences \cite{superI}, the goals of which would  determine the course of human civilisation. Even those AIs that are less powerful could have considerable impacts on human society -- as indeed they already have.

To avoid potential disruption, intelligences must be aligned with human preferences , goals, and values. This is the `alignment problem', and its resolution is by no means easy. Human values are hard to define. Asking people about their preferences typically elicits `stated' preferences, which are inadequate to explain behaviour \cite{kroes1988stated} -- people often don't state what they really want. `Revealed' preferences, identified by regarding people's actions as sufficiently informative of preference \cite{samuelson1938note}, assume that people are rational decision-makers, yet people are often far from rational in their decisions \cite{thinking_fast_slow}. Humans can be influenced by `irrelevant' changes to the architecture of choice-making \cite{thaler2004save,vspecian2019precarious}, making revealed preferences likewise insufficient indicators of actual preference.

Knowledge of preferences is important to companies\footnote{
For corporations, serving unmodified revealed preferences can result in reduced long-term customer satisfaction, leading to profit-loss (operational risk), individual and societal harm (reputational risk), and legal battles involving civil suits and regulator lawsuits (compliance risk). Therefore, serving a customer's true -- rather than revealed -- preferences is in the long-term interest of a for-profit company.
} and other organisations, and has been extensively studied by those engaged in researching machine learning. Many attempts have been made to obtain useful, correct, and compact representations of preferences \cite{viappiani2014preference}, a pertinent example of which was the original `Netflix Prize' \cite{Bennett2007TheNP}. This was based on user ratings of DVDs by consumers -- stated preferences -- but Netflix ultimately found revealed preferences to be more useful for its business objectives, and employed user engagement, and eventually retention, for their recommendation system \cite{GomezUribe2015TheNR}. Yet even this approach is problematic, as many organisations have tried to correct preference-related problems in their recommendation systems and other algorithms one at a time, rather than addressing them systematically \cite{stray2021you}.

Preferences and irrationalities -- ineffective attempts to achieve those preferences -- together determine human behaviour, which is alternatively  referred to as human 'policy'. But the determination is one way only. Preferences cannot themselves be deduced from policy: the collection of human preferences, values, and irrationalities, is strictly more complex than human policy \cite{armstrong2018occam}. Extra `normative assumptions ' need to be added to allow an algorithm to deduce human values from human behaviour. Work is ongoing to try and resolve this challenge \cite{hadfield2017inverse}.

By learning human policy an AI can attain considerable power ; it is thus important  that AIs learn human preferences (and thus some level of alignment) before they achieve that power. Learning human preferences will also generally increase the algorithm's knowledge of human policy , and hence their power over humans. Knowing human irrationalities can be even more dangerous, as this allows the AI to exploit these irrationalities from the beginning. Also, knowing irrationalities means that the AI cannot learn human preferences without also learning human policy.

It may thus be necessary for an algorithm to know most or all human preferences  before it is deployed unrestricted in the world. This point will be illustrated by models of recommendation system algorithms that suggest videos for users, and by different behaviours of algorithms that are either fully ignorant, that know a user's preferences, that know a user's irrationalities, and that know the full user's policy.

Especially dangerous is an unaligned AI with grounded knowledge \cite{harnad1990symbol} of human preferences, irrationalities and policy. The AI can then connect the grounded knowledge to the features of the world as it `knows' them. Grounded knowledge can result in discontinuous jumps in power, so that relatively weak AI systems might suddenly become very influential (see the forthcoming paper by the same authors, \cite{sturebsymb}).

\section{Learning preferences, irrationalities, and policy}

\subsection{Preferences, policies, and planers}

It is difficult, verging on impossible, to directly program an algorithm to follow the preferences of a human or a group of humans. For ambiguously defined tasks, it has proven much more effective to have the algorithm learn these preferences from data \cite{halevy2009unreasonable}, in this instance human behaviour as we go about our lives, choosing certain options and avoiding others.

Paper demonstrates, however, that one cannot learn the preferences (or the irrationalities) of irrational agents just through knowing their behaviour or policy.

Research \cite{armstrong2018occam} demonstrates, however, that the preferences or irrationalities of irrational agents cannot be learned simply by knowing their behaviour or policy. In the notation of the cited paper, $R$ is the reward function, $p$ is the rational (or irrational) decision module (called the `planner'), and $\pi$ is the agent's policy. The paper shows that the $(R, p)$ pair has strictly more information than $\pi$ does. If those three terms are seen as random variables (due to our uncertainty about them) and $H$ is information entropy\footnote{
A measure of the amount of uncertainty we have about the values, or, equivalently, how much information we gain upon knowing the values.
}, then
\begin{align*}
    H(R,p) > H(\pi).
\end{align*}

We take the most general position possible, and define irrationality as the deviation of the planner $p$ from a perfectly rational planner. We also posit that knowing human (ir)rationality and behaviour/policy permits the deduction of human preferences. Thus knowing $p$ and $\pi$ allows one to deduce $R$\footnote{
This is by no means a given for formal definitions of $p$ and $\pi$. But we are only excluding from consideration preferences that never make any difference to action in any conceivable circumstance.
}.

\subsection{Normative assumptions}

In order to infer $R$ (or $p$) from $\pi$, anyone seeking to program an algorithm would need to add extra `normative' assumptions, in order to bridge the difference between $H(\pi)$ and $H(R, p)$. Informally, we might say that it is impossible to learn human values unsupervised; it must be at least semi-supervised, with labeled data points being normative assumptions.

Some of these assumptions derive from shared properties of the human theory of mind (e.g., `if someone is red in the face and shouting insults at you, they are likely to be angry at you, and this is not a positive thing'), which in normal human experience appear so trivial that we might not even think it necessary to state them\footnote{
Indeed, such assumptions might be implicitly included in the code by programmers without them realising it, as they `correct obvious errors' or label data with `obvious' but value-laden labels.
}. Some might be regarded as `meta-preferences', pointing out how to resolve conflicts within the preferences of a given human (e.g., `moral values are more important than taste-based preferences'), or how to idealise human preferences into what a given person might want them to be (e.g., `remove any unconscious prejudices and fears within me'). Some might deal with how preferences should be extended to new and unexpected situations. Hand-crafting a full list of such assumptions would be prohibitively complex.

\subsection{Knowledge, power and AI alignment}

An AI is powerful if it knows how to affect the world to a great extent.  It is aligned if aims to maximise $R$ -- the values and preferences -- for all humans. Maximising $R$ requires that the AI knows it, of course, so alignment requires knowledge of $R$.

Generally speaking, knowing $\pi$ would make the AI more powerful, since it knows how humans would react and hence how best to manipulate them. And although knowing $\pi$ does not give $R$ and $p$ directly, the three are connected and so knowing $R$ or $p$, in whole or in part, would allow the AI to deduce much of $\pi$. Thus knowledge of human preferences leads to knowledge of human policy, and hence to potential power over humans.

Two scenarios are provided as examples where this is relevant: an aligned AI in development, and an unaligned AI of limited (constrained) power.

\subsubsection{Aligned AI in development}

Normative assumptions come in many different types, and humans are often not consciously aware of them. Thus AI programmers are unlikely to be able to code the whole set from first principles. Instead, they will experiment, trying out some assumptions, getting the AI to learn from human behaviour, seeing what the AI does, and refining the normative assumptions in an iterative loop.

Until this process is finished, the AI is unaligned: it is not fully motivated to maximise human preferences/rewards/values. If that AI becomes powerful during this intermediate stage there are likely to be consequences. It might be motivated to prevent its goals from being changed \cite{Omohundro:2008:BAD:1566174.1566226}, and attempt to prevent further normative assumptions from being added to it\footnote{
Some papers \cite{Orseau16} have demonstrated methods for combating this, but the methods are non-trivial to implement.
}. This outcome has been demonstrated in a paper \cite{2020arXiv200413654A} where algorithms that learn online (i.e., that learn their objectives while optimising these same objectives) are shown to have incentives to manipulate the learning process. A powerful AI with unaligned values could prove an existential risk to humanity \cite{superI}.

\subsubsection{Constrained unaligned AI}

A second type of unaligned AI is a constrained AI. This is an AI whose power is limited in some way, either through being `boxed' (constrained to only certain inputs and outputs \cite{oracleAI}), being a recommendation system, being only one agent in an economy of multiple agents (similar to how corporations and humans co-exist today), or simply being of limited power or intelligence.

However, the more such an AI knows about human policy, the more it can predict human reactions to its actions. So the more it knows, the more it is capable of manipulating humanity in order to gain power and influence, and to remove any constraints placed upon it.

\subsubsection{Knowing irrationalities, policies, and preferences}

Everything  else being equal, it is safest for both an aligned AIs in development and constrained AIs to know the maximum about $R$ (human preferences)  while knowing the minimum about $\pi$ (human policy) and $p$ (human irrationalities). Further, it is better that a constrained AI knows more about $R$, than about $\pi$, and more about $\pi$ than about $p$ (the worst-case scenario is if it only knows human irrationalities). The latter point arises from the position that if an AI knows only $R$ it can (and must) offer a decent trade in exchange for achieving its own goals, while one that knows only $p$ can (and must) exploit human irrationalities for its own purposes. This will be illustrated in the next section.

The second point comes from the fact that an AI that knows only $R$ can (and must) offer a decent deal in exchange for achieving its own goals, while one that knows only $p$ can (and must) only exploit our irrationalities for its purposes.

\section{Exploiting irrationalities vs. satisfying preferences}

Consider the following model: a constrained AI is a recommendation system that selects a daily video (for a website or an app). The system's goal is to cause the human user watch the video in full before they move on to something else\footnote{
Significantly, but typically, the goals of this recommendation system would be aligned neither with the users (who value their time and enjoyment) nor their parent company (who would value long-term retention and user-spending, or that users watch advertising, rather than that they watch a single video).
}. To do so, it selects one video from a collection of a $1,000$ daily topical video options each day, generated randomly.

In this model, each video has ten features, five related to human preferences (genre, storyline, characters, etc.) and five related to irrationalities (use of cliff-hangers, listicles, sound inconsistencies, etc.). For each video $v$ the recommendation system is given a timeline of the varying importance of each feature over the course of the video. Based upon this information, the system will average the information, characterising $v$ by two five-dimensional vectors: the preference vector $\overline{v_R}$ (each preference feature denoted by a number between 0 and 1) and the irrationality vector $\overline{v_p}$ (each irrationality feature also denoted by a number between 0 and 1).

Each user $h$ also has a collection of five preferences, $\overline{h_R}$ (which denote how much they enjoy certain aspects of the video) and five irrationality features $\overline{h_p}$ (which denote how susceptible they are to the video's `tricks'). These ten numbers also take values between $0$ and $1$.

Define $\Delta_R$ as the Euclidean distance between $h_R$ and $v_R$ (i.e. the Euclidean norm of $h_R-v_R$). Similarly, define $\Delta_p$ as the Euclidean distance between $h_p$ and $v_p$. Then the probability of the user watching the video in full is:
\begin{align*}
    e^{-\Delta_R^2-\Delta_p^2}.
\end{align*}
The recommendation system interacts with the same user each day, selecting a new video from the $1,000$ topical videos of that day. It knows how long the user watched previous videos, receiving a reward of $1$ whenever that length is the full length of the video (and $0$ otherwise).

This is formally known as a multi-armed contextual bandit online learning problem \cite{lu2010contextual}. Here the AI follows a greedy strategy: at each stage it selects the video that is most likely to be watched. To do so it uses a Monte Carlo simulation: generating a thousand random possible users, and computing their posterior probability of being $h$ by updating, on past observations, what videos $h$ watched and didn't watch. It then computes the probability of each random user watching a topical video on a given day, and calculates a weighted sum across the random users to get a final probability estimate for a given video being watched\footnote{
In practice, since we're only interested in one probability being higher than another, there is no need to renormalise the probabilities so that they sum to one.
}. It then selects the topical video with the highest probability of being watched in full.

\begin{figure}[!t]
\centering
\includegraphics[width=2.5in]{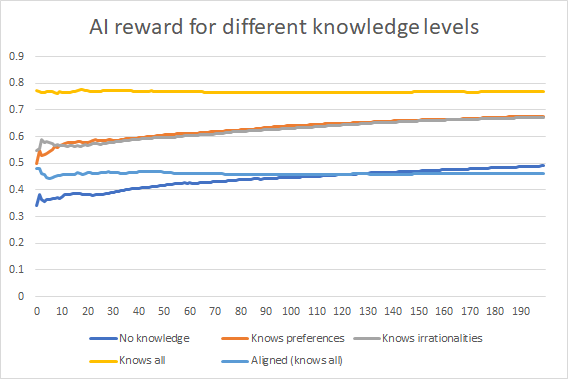}
\caption{Five recommendation systems select videos for users to watch. This graph plots the success of the systems that know nothing about the user (dark blue), know their preferences (orange), know their irrationalities (grey), or know both preferences and irrationalities (yellow). The light blue line is an aligned recommendation system that always chooses the video the user would prefer (though, significantly, because of user irrationalities, this is not necessarily the video the user is most likely to watch).}
\label{AI:reward}
\end{figure}

We consider four different possible systems: one that knows nothing of the user $h$ and has to learn from observing what they watch or don't, one that knows $\overline{h_R}$ (the user's preferences), one that knows $\overline{h_p}$ (the user's irrationalities), and one omniscient system that knows both (and therefore knows the human policy without needing to learn). For comparison, we also plot an aligned omniscient recommender system: this one knows the humans' preferences and selected the video that best fit with these. The results are computed for $150$ users, and then averaged; see \autoref{AI:reward}.%we have to choose between present and past tense. Present works so I have changed some bits for consistentcy. Future tense is not appropriate imo

The omniscient system convinces the user $h$ to watch the video roughly $77\%$  of the time. The `cold-start' system \cite{park2009pairwise}, which initially knows nothing, begins with less than $40\%$ success rate but this gradually increases as it learns more about the user. The systems that know preferences or irrationalities demonstrate performance levels between these two, and are equivalent to one another (due to the symmetry between preferences and irrationalities in this specific model). The aligned system only convinces the user to watch the video around $46\%$ of the time. This is because of user irrationalities: the video they'd most enjoy is not necessarily the one they are most likely to watch.

%I don't like the `get the user to' language here- using more technical language would be nice. 'Achieves a performance of x%' - though we can follow that from saying, 'from the perspective of the user, that means the user watches the video x% of the time'

Note that user $h$ only derives value from having their preferences satisfied, not from having their irrationalities exploited. Their reward needs to be inversely proportional to how closely the video matches their preferences, thus inversely proportional to $\Delta_R$.

Opportunity costs should be taken into account: if user h is not watching a video, then they would be doing some other activity that might be of value to them. Since reward functions are unchanged by adding constants, we choose to give a total reward of $0$ for these alternative activities. We set their reward for watching a video to be $10-100\Delta_R^2$.

\begin{figure}[!t]
\centering
\includegraphics[width=2.5in]{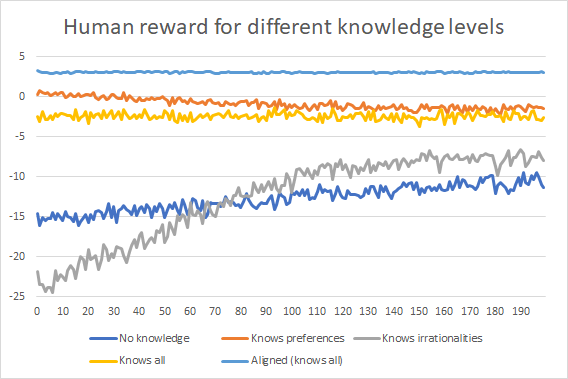}
\caption{Five recommendation systems select videos for users to watch, taking into account utility to the user (including opportunity costs). The systems may know nothing about the user (dark blue), know their preferences (orange), know their irrationalities (grey), or know both preferences and irrationalities (yellow). The light blue is an aligned recommendation system that always chooses the video that the user would prefer.}
\label{human:reward}
\end{figure}

So if $\Delta_R^2 < 1/10$, then watching the video is a net gain for the user: they derive more value from that activity than from doing anything else. If $\Delta_R^2 > 1/10$, then watching the video is a net negative: they would have been better served. The total human reward is graphed in \autoref{human:reward}.

Note that all non-aligned systems result in some disutility for their users: the opportunity cost removes any advantage in seeing a merely-adequate video. The system that knows only the user's preferences has the lowest disutility. It starts by offering videos that better align with the user's preferences until it learns their irrationalities as well, and user reward declines as the system selects less well-aligned videos that the user is nonetheless more likely to watch. The system that knows irrationalities exhibits the opposite behaviour. It starts by maximally exploiting irrationalities, then adds in more preference-aligned options as it learns user preferences, so that its disutility declines. The fully ignorant system has intermediate performance between the two.

As they learn, the behaviours of the non-aligned systems converge towards that of the omniscient system, which offers a consistent reward of around $-2.4$ (hence an overall disutility for the user). By contrast, the aligned system that chooses the best video provides a user reward of around $+3$ (hence an overall positive value for the user).

\subsection{Practical considerations}

The model presented above assumes that exploited irrationalities are of neutral value to humans, but the exploitation of irrationalities can have negative value in lived experience, including epistemic fragmentation \cite{milano2021epistemic}, preference amplification \cite{Kalimeris2021PreferenceAI}, and the distortion of human preferences caused by interaction with software agents \cite{burr2018analysis}. The human might also suffer disvalue from knowing or suspecting that their irrationalities are being exploited, and try to avoid this outcome.

One real-life experiment \cite{zhao2018explicit} that is similar to our model demonstrates the operation and negative value of exploited irrationalities. The authors of the study related clicks on hyperlinks with revealed preferences, and `human-in-the-loop' ratings with stated preferences. If clicks were true preferences, then maximising clicks (`optimising for engagement') would maximise value to the user. Yet the authors discovered something that they called `negative engagement', clicks made because the user had trouble finding the information they were looking for. A system optimising for engagement  would amplify this behaviour, negatively affecting user experience. This is a mistake of the algorithm designer rather than of the algorithm, which was merely following the instructions of the designer.

%In the context of our paper, these are represented by what we call `irrationality', although rather than constituting a cognitive bias of the human user, they represent the bias of a system that optimises for clicks ('engagement') in mistakenly considering these revealed preferences to be true preferences.

%Instead of the 'grounded knowledge' subsections below, I would put: 1) a 'conclusions' section. A dramatically shortened version of the 'grounded knowledge' sections below could become a short section called 'directions for further research', but I'd rather it be no more than a footnote. Why? Simplicity and succinctness. The paper is much more professional and likely to be cited and thought about if it is on a single topic, as this is not a review article or survey. (I know I'm not right about everything, but I'm confident I'm right about this.) 

\subsection{Grounded knowledge}

An algorithm has grounded knowledge when it has some symbolic data (academic publications, social media posts, user ratings, etc.) and a way of connecting that data with known elements of the world. For instance, a user might search for “Should I be worried my nose is still dripping from the fight last night?” The Google Flu Trends (GFT) web service \cite{dugas2013influenza} might flag this as evidence for influenza based on the `dripping nose' search terms, but it would not have done so if it understood the full meaning of the search phrase, which is clearly not flu related.

In the model above, the example algorithms are not exploiting the meaning of all the information they can access. There are ten features for each video, but they use only their average values; they also know how long the user watched a video, but only check whether this was full length or not.

A human with that information might deduce that a user is more likely to stop watching a video at the point where it is least pleasant to them -- the furthest from their preferences or irrationalities -- and could thus infer information about the user from that stopping point. It is less clear what the algorithm might have done if it `realised' what that extra information `meant'.

\begin{figure}[!t]
\centering
\includegraphics[width=2.5in]{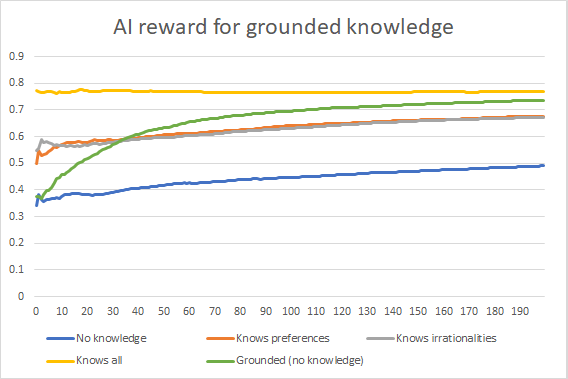}
\caption{The recommendation system exploiting grounded knowledge is shown in the green curve. Though it starts with the same performance as the fully ignorant system (blue), it quickly gains in performance, surpassing the systems that know preferences or irrationalities (orange and grey respectively) and converging towards the performance of the omniscient system that knows both the preferences and irrationalities of the user (yellow).}
\label{grounded:reward}
\end{figure}

It is however possible to model a grounded version of the algorithm. In our model, whenever the user rejects a video the system will obtain one piece of information about the preferences or irrationalities of that user\footnote{For this model, the algorithm will be given one of the ten preference or irrationality values at random, though this may be a value it already knows.}. Without modifying the rest of the algorithm, \autoref{grounded:reward} demonstrates how a grounded algorithm starts off as a poor recommendation system, comparable to the standard `no-knowledge' algorithm, but quickly achieves a level of performance comparable with the `omniscient' one.

\subsubsection{Grounded knowledge overhang: cached information}

One significant issue with grounded knowledge is that the algorithm might accumulate a sufficiently large collection of information and dramatically and discontinuously increase its power. For example, the `ignorant' algorithm of \autoref{grounded:reward} might suddenly `realise' the meaning of the information it has, and leap immediately from `no knowledge' to `grounded'. This might have a relatively trivial effect in a model such as a video recommendation system, but might have far greater effects for the recommendation systems currently in use that dominate real-world search results, news feeds, and social media.

\section{The potential severity of the problem}

Algorithms are typically designed with a measurable goal in mind, such as convincing a user to watch a video, click on a hyperlink, or re-subscribe to a service. Too little attention is paid to how the algorithm achieves that goal, or what information it uses to achieve it.

People are often on the lookout for use of sensitive personal information -- things like race, gender, sexuality, or medical information. This paper demonstrates that it is also dangerous for an algorithm to learn too much about human irrationalities. This applies no matter what the power of the algorithm; indeed, knowing too much about human irrationalities increases its effective power.

If an algorithm makes a discontinuous leap to a smarter and more powerful system, and `realises' that the knowledge inferred from the data it processes can be utilised for unaligned goals, then there is great potential risk for humanity. The data presented here shows that the risks of an algorithm manipulating easily-exploited irrationalities are greater than those that focus on preferences. This risk also applies to less powerful algorithms; knowing too much about human irrationalities will increase its effective power level.

The worst possible outcome would be where irrationalities are very easy to exploit, and where it is easy to deduce policy from preferences but hard to deduce preferences from policy. Constrained AIs would be the most powerful and exploitative, and in-development AIs would acquire a lot of power before they start to become even approximately aligned. By identifying this weakness in algorithm design, we can put in place checks and balances that limit the possibility of unaligned AIs becoming dangerous in this way.

%I don't like the following paragraph as it frames our paper weakly
%The toy example of this paper is a mild one, showing minor problems. But there is the potential for the problem being much more severe. The `omniscient' recommender system might exploit irrationalities to give the human user much greater disutility than in this example. Humans throughout history have already caused great pain and suffering by exploiting the irrationalities of others, and AIs might be much more effective.

%Thus the dynamics of how human policy, preferences, and irrationalities interact is worth more thorough analysis, of which this paper is but the beginning.%this is also a weakening paragraph

\section{Acknowledgments}
We wish to thank Nick Bostrom, Ryan Carey, Paul Christiano, Michael Cohen, Oliver Daniel-Koch, Matt Davis, Owain Evans, Tom Everrit, Adam Gleave, Tristan Harris, Ben Pace, Shane Legg, Laurent Orseau, Gareth Roberts, Phil Rosedale, Stuart Russell, Anders Sandberg, and Tanya Singh Kasewa, among many others.
This work was supported by the Alexander Tamas programme on AI safety research, the Leverhulme Trust, the Berkeley Existential Risk Institute, and the Machine Intelligence Research Institute.

\bibliographystyle{IEEEtran}

\bibliography{ref}

\end{document}